\documentclass[a4paper, 10pt, conference]{ieeeconf}      

\usepackage{graphicx} 
\usepackage[ruled,vlined]{algorithm2e}
\usepackage{subcaption}

\title{
Implementation of Q Learning and Deep Q Network For Controlling a Self Balancing Robot Model
}

\author{MD Muhaimin Rahman$^{1,a}$, SM Hasanur Rashid$^{2,a}$ and M.M Hossain $^{b}$\\
$^a$ Department of Mechanical Engineering, $^b$Department of Electrical and Electronic Engineering\\
Bangladesh University of Engineering \& Technology\\
Email:$^1$sezan92@gmail.com\\
$^2$hrshovon@gmail.com\\
$^3$monir.eee.buet@gmail.com\\
}

\begin{document}

\maketitle
\thispagestyle{empty}
\pagestyle{empty}

\begin{abstract}
In this paper, the implementation of two Reinforcement learnings namely, Q Learning and Deep Q Network(DQN) on a Self Balancing Robot Gazebo model has been discussed.  The goal of the experiments is to make the robot model learn the best actions for staying balanced in an environment. The more time it can stay within a specified limit , the more reward it accumulates and hence more balanced it is. Different experiments with different learning parameters on Q Learning and DQN are conducted and the plots of the experiments are shown.

\end{abstract}

\section{INTRODUCTION}
Control systems is one of the most important aspects of Robotics Research. Gazebo is one of the most robust multi robot simulators at present. The ability to use Robot Operating System (ROS) with Gazebo makes it more powerful. But there are very few documentations on how to use ROS and Gazebo for Controllers development. In our previous paper, \cite{icmePaper}, we attempted to demonstrate and document the use of PID, Fuzzy logic and LQR controllers using ROS and Gazebo on a self balancing robot model. Later on, we have worked on Reinforcement learning. In this paper,  implementation of Q Learning and Deep Q Network on the same model is discussed.
The paper is structured as follows . Section \ref{Related} shows the related works on the subject. Section \ref{Model} discusses the Robot Model . Section \ref{RL} shows the Implementation of Q Learning and DQN as controllers . Finally, section \ref{conc} is the conclusion.
\section{Related Works}
\label{Related}
Lei Tai and Ming Liu \cite{cnnRL}, had worked on Mobile Robots Exploration using CNN based reinforcement learning. They trained and developed turtlebot Gazebo simulation to develop an exploration strategy based on raw sensor value from RGB-D sensor. The company \textit{ErleRobotics} have extended OpenAI environment to Gazebo \cite{gym-gazebo}. They have deployed Q-learning and Sarsa algorithms for various exploratory environments. Loc Tran et al \cite{nasaRL} developed training model for an Unmanned aerial vehicle to explore with static obstacles  in both Gazebo and real world. Their proposed Reinforcement learning is unclear from the paper. Volodymyr Sereda \cite{volodRL} used Q - learning on a custom Gazebo model using ROS in  for exploration strategy. Rowan Border \cite{RowanRL} used Q-learning with neural network presentation for robo search and rescue using ROS and Turtlebot.
\section{Robot Model}
\label{Model}
\begin{figure}

\includegraphics[width=\linewidth]{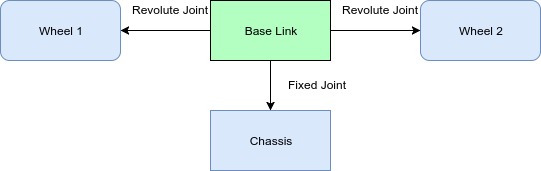}
\caption{Simple Block Diagram of the Model}
\label{BlockDiagram}
\end{figure}
\begin{figure}[b]

\includegraphics[width=\linewidth]{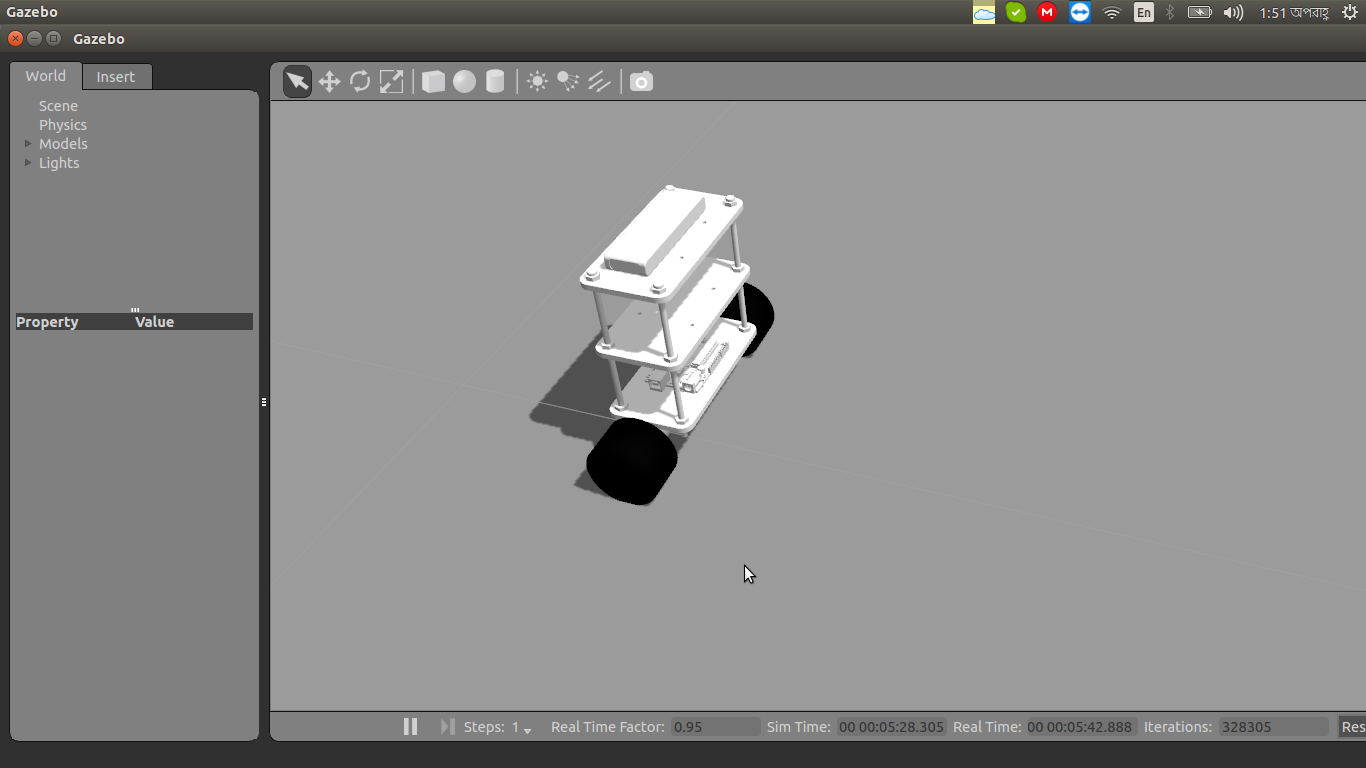}
\caption{Gazebo Model}
\label{GazeboModel}
\end{figure}

The Robot Model is described in paper \cite{icmePaper}. It has one chassis and two wheels. The task of the model is to keep the robot balanced i.e. keeping it's pitch angle in between $\pm 5^0$. The more it remains in between the limits , the more it gets the reward. The Fig. \ref{BlockDiagram} shows the block diagram and the Fig.\ref{GazeboModel} shows the Gazebo model.

\subsection{Controller}
\begin{figure}[t]
\includegraphics[width=\linewidth]{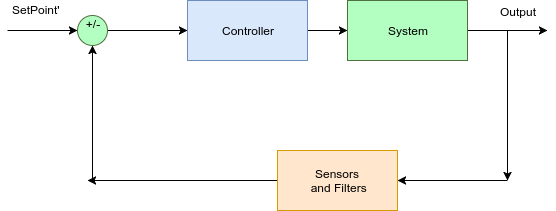}
\caption{Controller Block Diagram}
\label{Controller}
\end{figure}

The robot's IMU sensor measures the roll, pitch and yaw angles of the chassis every second and sends it to the controller. The controller calculates optimum action value to make the chassis tilt according to set point.
 Fig. \ref{Controller} shows the control system of the robot.

\section{Reinforcement Learning Methods as Controllers}
\label{RL}
In \cite{icmePaper}, we worked on traditional Controllers like PID, Fuzzy PD,PD+I \& LQR . The biggest problem with those methods is that , they need to be tuned manually. So, reaching optimal values of Controllers depends on many trials and errors. Many a times optimum values aren't reached at all. The biggest benefit of Reinforcement learning algorithms as Controllers is that , the model tunes itself to reach the Optimum values. The following two sections discuss Q Learning and Deep Q Network.
\subsection{Q Learning}
Q- learning was developed by Christopher John Cornish Hellaby Watkins \cite{QLearning}. According to Watkins, "it provides agents with the capability of learning to act optimally in Markovian domains by experiencing the consequences of actions, without requiring them to build maps of the domains." \cite{QLearning2}. In a Markovian domain, $Q$ function- the model to be generated using the algorithm- calculates the expected utility for a given finite state $s$ and every possible finite action $a$. The agent - which is the robot in this case- selects the optimum action $a$ having the highest value of $Q(s,a)$ , this action choosing rule is also called Policy. \cite{QLearning2} . Initially, the $Q(s,a)$ function values are assumed to be zero. After every training step , the values are updated according to the following equation
\begin{equation}
 Q(s,a_t) \gets  Q(s,a_t)+ \alpha(r+\gamma maxQ(s_{t+1},a)) 
\label{QEquation}
\end{equation}
\subsubsection{Algorithm}
The objective for the Model in our project is to keep it within limits i.e. $\pm 5^0$ . At first , the robot model, $Q$ matrix, Policy $\pi$ are initialized . There are some interesting points to make. The states are not finite. Within limit range , hundreds and thousands of pitch angles are possible. Having thousands of columns is not possible. So, the state values were discretized. We discretized the values to 20 discrete state angles from $-10^0$ to $10^0$. For action value, we chose 10 different velocities. They are $[-200,-100,-50,-25,-10,10,25,50,100,200] ms^{-1}$. The $Q$ matrix had 20 columns , each column  representing a state and 10 rows each representing every action. Initially ,the Q -values were assumed to be $0$ and random actions were specified for every state in the policy $\pi$ . The training was done for 1500 episodes and in each episode, the training was iterated $2000$ times . At the beginning of each episode, the simulation was refreshed. Whenever the robot's state exceeded the limit it was penalized by assigning reward to $-100$ . The $Q$ Table is updated at each step according to equation \ref{QEquation}. The Algorithm \ref{QLearningAlgorithm} shows the full algorithm.  
\subsubsection{Result and Discussion}
\begin{figure}
\includegraphics[width=\linewidth]{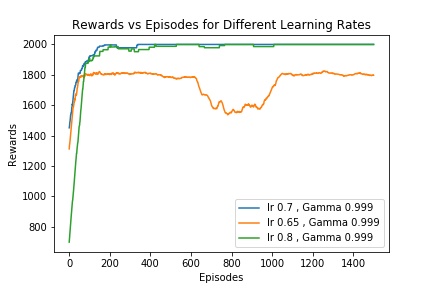}
\caption{Rewards for different $\alpha$}
\label{fig:Q_Comparison}
\end{figure}

The simulation was run for three different $\alpha$ values $(0.7 , 0.65 , 0.8)$ , with $\gamma$ value of  $(0.999) $. The Fig. \ref{fig:Q_Comparison} shows the Rewards vs Episodes for those $\alpha$s. It is evident that, the robot couldn't reach the target rewards within the training period for those learning rates. We see that, for the $\alpha$ values $0.7$ and $0.8$, the robot reaches maximum possible accumulated rewards, 200, within 400 episodes. The curve with $\alpha$ value $0.7$ is less stable compared to that of $0.8$. But The curve with $\alpha$ value $0.65$ never reaches the maximum accumulated reward.
\begin{algorithm}
\SetAlgoLined
\caption{Q Learning Algorithm as applied in the system}
\label{QLearningAlgorithm}
Initialize Robot\;
Initialize Q Matrix $Q$\;
Initialize Policy $\pi$\;
Initialize Penalty Reward $pen$\;
\For{number of episodes}{
Reset simulation \;
Wait for 1 second \;
Pause simulation \;
Read the pitch angle $\phi$ of the robot \;
$state \gets \phi $ \;
Unpause simulation \;
\For{number of iterations}{
Generate a random number $rand$\;
\If{ $ rand \leq \epsilon$ }{
take random action \;
}
\Else{
take action based on $\pi$ \;}
$state_{new} \gets \phi $\;
Pause simulation\;
\If{ absolute value of $state_{new} \geq limit$}{
\If{ $reward_{total} \leq Target$}{
$reward \gets pen $\;
Update $Q$ \;
Update $\pi$\;}
Break \;

\Else{
Print Passed \;
Break  \;
}
}
\Else{
$reward \gets$ 1\;
Update $Q$\;
Update $\pi$
$state \gets state_{new}$
}
}

}
\end{algorithm}
\subsection{Deep Q Network (DQN)}
V Mnih et al \cite{DQNPaper} first used Deep Learning as a variant of Q Learning algorithm to play six games of  Atari 2600 , which outperformed all other previous algorithms. 
. In their paper, two unique approaches were used. 
\begin{itemize}
\item{Experience Replay}
\item{Derivation of Q Values in one forward pass}
\end{itemize}

\subsubsection{Experience Replay}
The technique of Experience Replay,  experiences of an agent , i.e. $(state,reward,action,state_{new})$ are stored over many episodes . In the learning period, after each episode random batches of data from experience are used to update the model. \cite{DQNPaper}. There are several benefits of such approach. According to the paper,
\begin{itemize}
\item{It allows greater data efficiency as each step of experience can be used in many weight updates}
\item{Randomizing batches breaks correlations between samples}
\item{Behaviour distribution is averaged over many of its previous states}
\end{itemize}
\subsubsection{ Derivation of Q Values in one Forward Pass}
\begin{figure}[t]
\includegraphics[width=\linewidth]{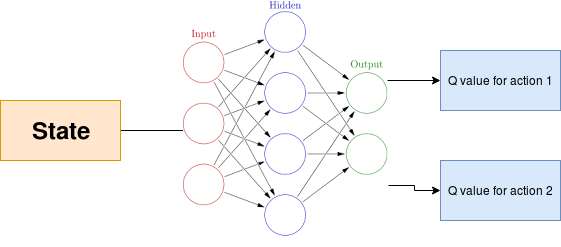}
\caption{ Sample Deep Q Network Architecture}
\label{fig:DQN}
\end{figure}

In the classical Q learning approach, one has to give state and action as an input  resulting in $Q$  value for that state and action. Replicating this approach in Neural Network is problematic. Because in that case one has to give state and action for each possible action of the agent to the Model. It will lead to many forward passes in the same model. Instead , they designed the model in such a way that it will predict Q values for each action for a given state . As a result, only one forward pass is required. Figure \ref{fig:DQN} shows a sample architecture for one state with two actions 

\subsubsection{Implementation on the Robot Model}
\begin{algorithm}
\SetAlgoLined
\caption{DQN Algorithm as applied in the system}
\label{DQNAlgorithm}
Initialize Robot\;
Initialize model $M$\;
Initialize Penalty Reward $pen$\;
\For{number of episodes}{
Reset simulation \;
Wait for 1 second \;
Pause simulation \;
Read the pitch angle $\phi$ of the robot \;
$state \gets \phi $ \;
Unpause simulation \;
\For{number of iterations}{
Generate a random number $rand$\;
\If{ $ rand \leq \epsilon$ }{
take random action \;
}
\Else{
$Q \gets M(state)$\;
 $action \gets action for max(Q)$\; }
$state_{new} \gets \phi $\;
Pause simulation\;
\If{ absolute value of $state_{new} \geq limit$}{
\If{ $reward_{total} \leq Target$}{
$reward \gets pen $\;
$experience \gets (state,reward,action,state_{new})$\;
Add Experience to Memory\;}
Break \;

\Else{
Print Passed \;
Break  \;
}
}
\Else{
$reward \gets$ 1\;
$experience \gets (state,reward,action,state_{new})$\;
Add Experience to Memory\;
$state \gets state_{new}$
}
}
Take random minibatch of Experience\;
\If{$reward == pen$}{
$Q_{pred} \gets reward$\;}
\Else{
$Q_{pred} \gets reward + \gamma max(Q(state_{new},action))$}\;
Train the model according to loss $abs(Q_{pred}(state,action)-Q_{pred}(state,action))$\;
}
\end{algorithm}
The implementation of the DQN on our Robot model is similar to $Q$ Learning Method. However, there are some exceptions. At first, a  model was initialized instead of Initializing $Q$ matrix. In the $\epsilon$ greedy policy , instead of choosing action based on policy $\pi$ , $Q$ values were calculated according to the model. At the end of every episode, the model was trained using random minibatches of Experience. At first, an architecture with 2 hidden Relu layers of 20 units was selected. The last layer was a Linear Dense layer with 10 units. With the $\gamma$ 0.999 and $\alpha$ of $(0.65,0.7,0.8)$ .  Algorithm \ref{DQNAlgorithm} shows the DQN algorithm as implemented on the robot model.

\subsubsection{Result and Discussion}
\begin{figure}
\includegraphics[width =\linewidth]{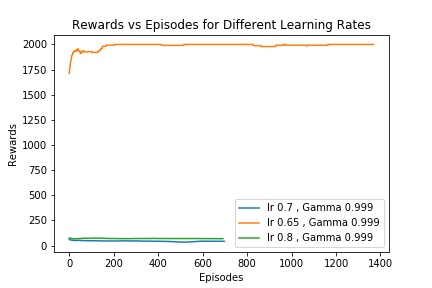}
\caption{"Rewards for three different $\alpha$s with $\gamma$ 0.999"}
\label{DQN1}
\end{figure}

\begin{figure}
\includegraphics[width =\linewidth]{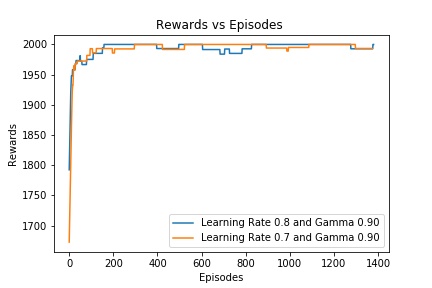}
\caption{Rewards vs Episodes for New Architecture}
\label{DQN2}
\end{figure}

From figure \ref{DQN1}, we see that the total rewards for $\alpha$ $0.65$ is significantly higher. It starts approximately from 1750 and reaches the maximum total rewards, 2000 within 200th episode. But the accumulated rewards  with $\alpha$ values of $0.7 $and $0.8$ are very low. They have accumulated rewards approximately 50-60  for the whole time. Later , the architecture was changed to 2 hidden layers of 40 Relu Units . The $\gamma$ was selected to be $0.9$ . Figure \ref{DQN2} shows that the both curves reached highest accumulated rewards within 200 episodes in the new configuration.

\section{Comparison to traditional methods}
\begin{figure}
\includegraphics[width=\linewidth]{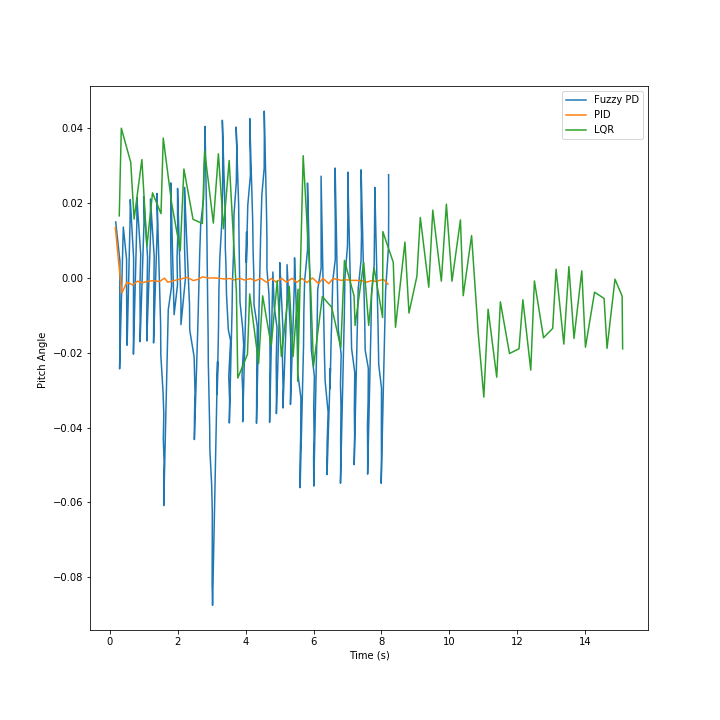}
\caption{Performance curve for PID, Fuzzy logic and LQR}
\label{Comparison}
\end{figure}

In our previous paper, \cite{icmePaper} , we compared PID, Fuzzy Logic and LQR . The Fig. \ref{Comparison} shows the performance curves for different controllers. It shows that LQR and Fuzzy controllers were not so stable like PID. Although we had to tune all of them manually. But the Fig. \ref{DQN2} shows that DQN learning is more stable after some iterations.
\section{CONCLUSION and FUTURE WORK}
\label{conc}
The implementation of  $Q $ Learning and Deep $Q$ Network as a controller in the Gazebo Robot Model was shown in this paper. It showed the details of the algorithms. However, some further improvments can be done. Like, It was assumed that the robot will work on Markovian State space , which generally not the case. In general Inverted pendulum models are Non-markovian models . So there must exist some kind of dependancies among the states. So In future, Recurrent Neural Network have a great possibility. Moreover,  10 predefined values of velocities for action were used. In the real world application, action values have continous range. So for more complex models, this method may not work. In that case, deep reinforcment learning algorithms with continuous action space like Actro Critic Reinforcement Learning algorithm \cite{ActorCritic} can be used.  Finally, this work should be improved for real world scenarios.
\addtolength{\textheight}{-12cm}   
\section{Declarations}
\subsection{Competing Interests}
The Authors declare that they have no competing interests
\subsection{Author's contribution}
The original project is this paper and \cite{icmePaper} . The Contributions of MD Muhaimin Rahman is the simulations and writing of this paper. The contributions of SM Hasanur Rashid and M.M. Hossain is reviewing both papers.
\subsection{Funding}
The paper has no external source of funding.
\bibliographystyle{IEEEtran}
\bibliography{bibliography}

\begin{thebibliography}{10}
\providecommand{\url}[1]{#1}
\csname url@rmstyle\endcsname
\providecommand{\newblock}{\relax}
\providecommand{\bibinfo}[2]{#2}
\providecommand\BIBentrySTDinterwordspacing{\spaceskip=0pt\relax}
\providecommand\BIBentryALTinterwordstretchfactor{4}
\providecommand\BIBentryALTinterwordspacing{\spaceskip=\fontdimen2\font plus
\BIBentryALTinterwordstretchfactor\fontdimen3\font minus
  \fontdimen4\font\relax}
\providecommand\BIBforeignlanguage[2]{{%
\expandafter\ifx\csname l@#1\endcsname\relax
\typeout{** WARNING: IEEEtran.bst: No hyphenation pattern has been}%
\typeout{** loaded for the language `#1'. Using the pattern for}%
\typeout{** the default language instead.}%
\else
\language=\csname l@#1\endcsname
\fi
#2}}

\bibitem{icmePaper}
\BIBentryALTinterwordspacing
M.~D.~M. Rahman, S.~M.~H. Rashid, K.~M.~R. Hassan, and M.~M. Hossain,
  ``Comparison of different control theories on a two wheeled self balancing
  robot,'' \emph{AIP Conference Proceedings}, vol. 1980, no.~1, p. 060005,
  2018. [Online]. Available:
  \url{https://aip.scitation.org/doi/abs/10.1063/1.5044373}
\BIBentrySTDinterwordspacing

\bibitem{cnnRL}
\BIBentryALTinterwordspacing
L.~Tai and M.~Liu, ``Mobile robots exploration through cnn-based reinforcement
  learning,'' \emph{Robotics and Biomimetics}, vol.~3, no.~1, p.~24, Dec 2016.
  [Online]. Available: \url{https://doi.org/10.1186/s40638-016-0055-x}
\BIBentrySTDinterwordspacing

\bibitem{gym-gazebo}
\BIBentryALTinterwordspacing
I.~Zamora, N.~G. Lopez, V.~M. Vilches, and A.~H. Cordero, ``Extending the
  openai gym for robotics: a toolkit for reinforcement learning using {ROS} and
  gazebo,'' \emph{CoRR}, vol. abs/1608.05742, 2016. [Online]. Available:
  \url{http://arxiv.org/abs/1608.05742}
\BIBentrySTDinterwordspacing

\bibitem{nasaRL}
\BIBentryALTinterwordspacing
L.~D. Tran, C.~D. Cross, M.~A. Motter, J.~H. Neilan, G.~Qualls, P.~M. Rothhaar,
  A.~Trujillo, and B.~D. Allen, ``Reinforcement learning with autonomous small
  unmanned aerial vehicles in cluttered environments,'' \emph{15th AIAA
  Aviation Technology, Integration, and Operations Conference}, Jun 2015.
  [Online]. Available: \url{https://doi.org/10.2514/6.2015-2899}
\BIBentrySTDinterwordspacing

\bibitem{volodRL}
V.~Sereda, ``Machine learning for robots with ros,'' Undergraduate thesis,
  Maynooth University, Maynooth, Co. Kidare, 2017.

\bibitem{RowanRL}
R.~Border, ``Learning to save lives: Using reinforcement learning with
  environment features for efficient robot search,'' White Paper, University of
  Oxford, 2015.

\bibitem{QLearning}
C.~J. Watkins, ``Learning from delayed rewards,'' Ph.D. dissertation, Kings's
  Collenge, London, May 1989.

\bibitem{QLearning2}
\BIBentryALTinterwordspacing
C.~J. C.~H. Watkins and P.~Dayan, ``Q-learning,'' \emph{Machine Learning},
  vol.~8, no.~3, pp. 279--292, May 1992. [Online]. Available:
  \url{https://doi.org/10.1007/BF00992698}
\BIBentrySTDinterwordspacing

\bibitem{DQNPaper}
\BIBentryALTinterwordspacing
V.~Mnih, K.~Kavukcuoglu, D.~Silver, A.~Graves, I.~Antonoglou, D.~Wierstra, and
  M.~A. Riedmiller, ``Playing atari with deep reinforcement learning,''
  \emph{CoRR}, vol. abs/1312.5602, 2013. [Online]. Available:
  \url{http://arxiv.org/abs/1312.5602}
\BIBentrySTDinterwordspacing

\bibitem{ActorCritic}
\BIBentryALTinterwordspacing
V.~Mnih, A.~P. Badia, M.~Mirza, A.~Graves, T.~Lillicrap, T.~Harley, D.~Silver,
  and K.~Kavukcuoglu, ``Asynchronous methods for deep reinforcement learning,''
  in \emph{Proceedings of The 33rd International Conference on Machine
  Learning}, ser. Proceedings of Machine Learning Research, M.~F. Balcan and
  K.~Q. Weinberger, Eds., vol.~48.\hskip 1em plus 0.5em minus 0.4em\relax New
  York, New York, USA: PMLR, 20--22 Jun 2016, pp. 1928--1937. [Online].
  Available: \url{http://proceedings.mlr.press/v48/mniha16.html}
\BIBentrySTDinterwordspacing

\end{thebibliography}

\end{document}